\pdfoutput=1

\documentclass[11pt]{article}

\usepackage[]{acl}
\usepackage{graphicx}

\usepackage{times}
\usepackage{latexsym}
\usepackage{subcaption}
\usepackage{arydshln}

\usepackage[T1]{fontenc}

\usepackage[utf8]{inputenc}

\usepackage{microtype}
\usepackage{amsmath}
\usepackage{multirow}
\usepackage{hyperref}
\usepackage{todonotes}
\usepackage{cleveref}

\crefname{section}{§}{§§}

\title{Spurious Correlations in Reference-Free Evaluation of Text Generation}

\author{
Esin Durmus$^1$\Thanks{  Equal contribution.}~\;~ Faisal Ladhak$^2$\footnotemark[1]~\;~ Tatsunori Hashimoto$^1$\\
$^1$Stanford University~\;~ $^2$Columbia University\\
\texttt{\href{mailto:esindurmus@cs.stanford.edu}{esindurmus@cs.stanford.edu}}~\;~\texttt{\href{mailto:faisal@cs.columbia.edu}{faisal@cs.columbia.edu}} \\
\texttt{\href{mailto:thashim@stanford.edu}{thashim@stanford.edu}}
}

\begin{document}
\maketitle

\begin{abstract}
Model-based, reference-free evaluation metrics have been proposed as a fast and cost-effective approach to evaluate Natural Language Generation (NLG) systems. 
Despite promising recent results, we find evidence that reference-free evaluation metrics of summarization and dialog generation may be relying on spurious correlations with measures such as word overlap, perplexity, and length. We further observe that for text summarization, these metrics have high error rates when ranking current state-of-the-art abstractive summarization systems.
We demonstrate that these errors can be mitigated by explicitly designing evaluation metrics to avoid spurious features in reference-free evaluation.
\end{abstract}

\section{Introduction}

Building reliable automated evaluation metrics is a key factor for quick development
of better NLG systems. Recent work has proposed reference-free evaluation metrics as a way to judge the quality of generated outputs without the need for human references \cite{DBLP:journals/corr/abs-2006-14799}. Many of these reference-free evaluations achieve remarkably high correlations with human evaluations, raising hopes that they may soon become a viable alternative to expensive human evaluations \cite{kryscinski-etal-2020-evaluating, goyal-durrett-2020-evaluating, sinha-etal-2020-learning, phy-etal-2020-deconstruct, gao-etal-2020-dialogue}. 

However, simply looking at correlation with human scores may not be sufficient to determine the efficacy and robustness of an evaluation metric. In our work, we study recently proposed reference-free evaluation metrics of text summarization and dialog generation. We find that it is possible to achieve similar levels of correlation with human judgment, using simple spurious correlates such as word overlap, length, and perplexity. Furthermore, we find that the learned metrics have a relatively high correlation with the spurious correlates as compared to human scores, which suggests that these metrics may rely heavily on spurious correlations. This may be a potential explanation for the robustness issues that are observed in recent work, despite the seemingly high reported correlations with human judgements \cite{gabriel-etal-2021-go, yeh-etal-2021-comprehensive}.

We further analyze reference-free faithfulness evaluation metrics and show that the reliance on spurious correlations leads to errors in model selection and development. First, we show that word overlap, a spurious correlate for the task, does as well as recently proposed reference-free metrics at system-level ranking. Then, we look at rankings amongst systems that are relatively abstractive and faithful, i.e., the current state of the art, and find that these learned metrics perform significantly worse for these systems. This is because word-overlap is not a good measure for ranking these systems in terms of their faithfulness since all of these systems have similarly low word overlap. This suggests that we need metrics that are not overly reliant on word overlap in their faithfulness prediction.

Finally, we explore whether a simple mitigation strategy of adversarially training a faithfulness evaluation metric to avoid spurious correlates can lead to a more robust metric. We find that our adversarially trained metric performs well at overall pairwise ranking while having a significantly lower correlation with the spurious correlate of word-overlap. Crucially, we show that our proposed metric has improved performance in ranking between abstractive and faithful systems, which is a failure mode for existing reference-free faithfulness evaluation metrics.

\section{Reference-free Evaluation of Text Generation}
We begin by defining the task of reference-free evaluation, as well as the \emph{example-level} and \emph{systems-level} evaluation of these metrics.

We define a reference-free evaluation metric as a function $F(x,y)$ that can assign a quality score to an output sequence $y$ for a given input sequence $x$. The goal of a reference-free evaluation metric $F(x,y)$ is to assign high scores to desirable outputs $y$ for some attribute, such as the faithfulness of a summary. Measuring the quality of this metric is challenging, and prior work has relied upon correlation to human judgments $H(x,y)$. 

\textbf{Example-level evaluation:} A number of existing reference free evaluations rely upon a procedure which we call \emph{example-level} human correlations \cite{fabbri2020summeval, phy-etal-2020-deconstruct, sinha-etal-2020-learning}, which measures the effectiveness of a metric by computing a Pearson or Spearman correlation $\text{corr}_{p_{\text{eval}}}(H(x, y),F(x, y))$ over some sampled evaluation data $p_{\text{eval}}(x,y)$.

\textbf{System-level evaluation:} An alternative approach to evaluation is \emph{systems-level} rankings \cite{mathur-etal-2020-tangled, DBLP:journals/corr/abs-2107-10821}, which we define as the ability to identify which model is better amongst a set of models $M$. $F$ is evaluated via its accuracy in matching human evaluation $H$ on all pairs $(m_i, m_j) \in M\times M$ where $m_i \neq m_j$. 

The definitions of example and system level correlations suggest that evaluations of these metrics may have a strong dependence on the example and systems distributions $p_{\text{eval}}(x,y)$ and $M$. As an example, consider an evaluation for dialogue response quality. Building a truly accurate predictor for dialogue response quality is challenging, but if $p_{\text{eval}}(x,y)$ consists of all either professionally written examples or ungrammatical nonsense, a simple grammar checker would perform exceedingly well.

This is an instance of what is called a spurious correlate. More formally, we define this as some attribute $S(x,y)$ which is correlated with $H$ in $p_{\text{eval}}(x,y)$ but is not correlated with $H$ for a carefully constructed test distribution $p_{\text{test}}(x,y)$. We say that $F$ is \textit{spuriously correlated} with $S$ if:
\begin{center}
\begin{enumerate}
    \item $F$ and $H$ are highly correlated under $p_{\text{eval}}(x,y)$ but not under $p_{\text{test}}(x,y)$.
    \item $F$ remains correlated with $S$ under $p_{\text{test}}(x,y)$.  
\end{enumerate}
\end{center}

\section{Example-level Analysis of Learned Evaluation Metrics}
\label{sec:example_level}

In this section, we look at example-level Spearman correlations with human judgements for reference-free evaluation metrics that have been proposed for summarization and dialog generation. We compare the metrics to spurious correlates such as word-overlap, length and perplexity, in order to understand  whether the metrics can perform better than these simple measures. We also measure to what extent the proposed metrics are correlated with these spurious measures.

\subsection{Faithfulness Evaluation in Text Summarization} \label{section_faithfulness_summarization}
State-of-the-art text summarization models are capable of producing fluent summaries. However, they suffer from generating information that is not consistent (i.e., unfaithful) with the information in the source article \cite{DBLP:conf/aaai/CaoWLL18}. Prior work showed that reference-based metrics are not able to capture such consistency errors \cite{falke-etal-2019-ranking}. This motivated researchers to build evaluation metrics to capture these faithfulness issues since collecting human evaluations for faithfulness is expensive and time-consuming \cite{wang-etal-2020-asking, durmus-etal-2020-feqa, kryscinski-etal-2020-evaluating, goyal-durrett-2020-evaluating}. 

In this section, we analyze recently proposed reference-free faithfulness evaluation metrics and compare their performance against the spurious correlate of word overlap. Furthermore, we analyze the correlation between the learned metrics and word overlap to understand to what extent these metrics rely on spurious correlations. We focus on learned entailment-based faithfulness evaluation metrics due to their high performance in identifying faithfulness issues \cite{pagnoni-etal-2021-understanding}. In particular we evaluate FactCC \cite{kryscinski-etal-2020-evaluating} and DAE \cite{goyal-durrett-2021-annotating}, which have been shown to achieve higher example-level correlations with human judgements than existing faithfulness evaluation metrics \cite{pagnoni-etal-2021-understanding}.

\textbf{FactCC.} \newcite{kryscinski-etal-2020-evaluating} proposed an entailment-based method where they train a BERT-based model to predict whether or not the source article entails a summary. To train this model, they generate synthetic training data by applying a set of transformations to source article sentences in order to get article, summary pairs. They evaluate their approach on the CNN/DM dataset \cite{see-etal-2017-get} and report a high accuracy on example-level comparisons on a human-annotated test set.

\textbf{DAE.} \newcite{goyal-durrett-2021-annotating} collected human annotations at the word-level and arc-level to study faithfulness at a finer granularity. They also trained a dependency arc entailment model for faithfulness detection \cite{goyal-durrett-2020-evaluating}. They evaluate on the same test set as \newcite{kryscinski-etal-2020-evaluating} and report improved results over FactCC.

We look at how these learned, reference-free metrics compare with word overlap -- a simple spurious correlate. One simple measure of whether a generated summary is faithful is to look at its word overlap with the source article; summaries with a higher word overlap are more likely to be faithful \cite{DBLP:journals/corr/abs-2108-13684}. However, this measure of faithfulness is spurious because it cannot distinguish between faithful and unfaithful summaries that have similar word overlap. In particular, we look at two metrics of word-overlap following \newcite{grusky-etal-2018-newsroom}: \textit{coverage} and \textit{density}. \textit{Coverage} measures the percentage of the words in the summary that are also present in the article. \textit{Density} instead looks at the average length of the segments in the summary that are extracted from the article.

\begin{figure}[]
\centering
\includegraphics[width=0.53\textwidth]{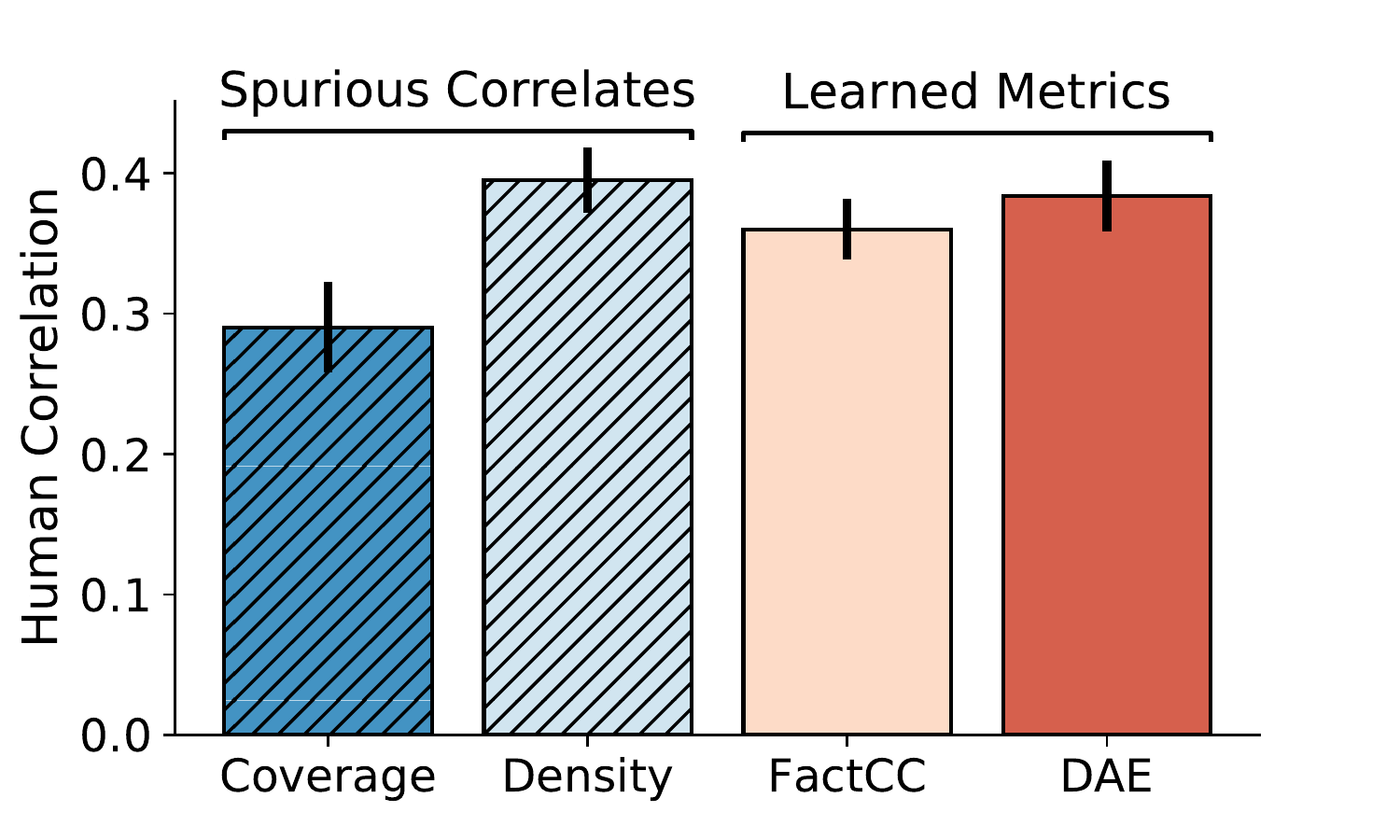}
\caption{Correlation of the spurious correlates and learned metrics with human scores. Density, a spurious correlate, achieves similar performance as DAE and performs significantly better than FactCC.}
\label{fig:sum_metric_human}
\end{figure}

\begin{table}[]
    \centering
    \begin{tabular}{|l|c|c|}
    \hline
      Metric & Human & Density \\ 
      \hline
      FactCC & 0.36 & \textbf{0.59} \\ 
       DAE & 0.38 & \textbf{0.76} \\   
       \hline
    \end{tabular}
    \caption{Correlation of FactCC and DAE scores with humans vs density. Both learned metrics have a significantly higher correlation with density than human scores.}
    \label{tab:sum_metric_correlate}
\end{table}

 \begin{figure*}
\centering
\includegraphics[width=\textwidth]{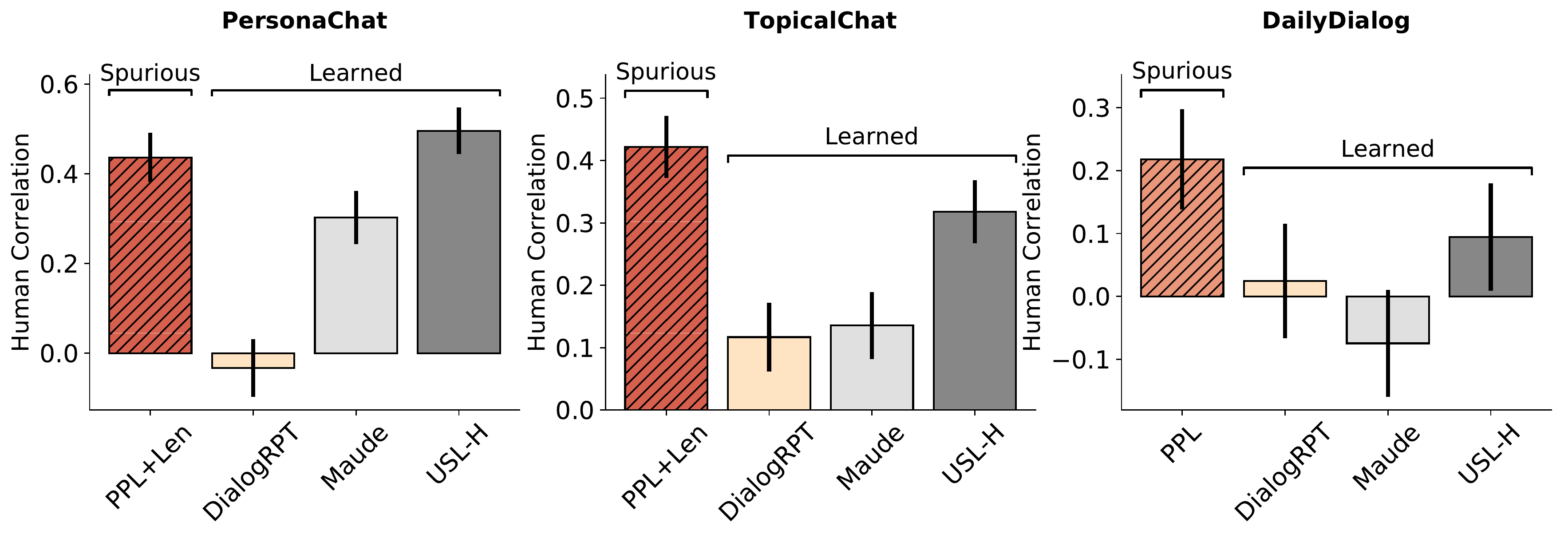}
\caption{Correlation of the spurious correlates and learned metrics with human scores. PPL+Len represents a simple combination of perplexity (PPL) and length features. The best spurious correlate performs significantly better than all learned metrics on TopicalChat, and performs similarly to the best learned metric on PersonaChat and DailyDialog.}
\label{fig:dialog_metric_human}
\end{figure*}

\textbf{Results.} We use the large-scale faithfulness human annotations collected by \newcite{fabbri2020summeval} for $16$ summarization models on the CNN/DM dataset \cite{see-etal-2017-get} for our analysis. Figure \ref{fig:sum_metric_human} shows the example-level correlations with human scores for each of the factuality metrics as well as the spurious correlates. We note that \textit{density} has a similar correlation with human scores as DAE, and is significanlty\footnote{All numbers reported in the paper are bootstrap means over $1000$ bootstrap samples. We use a one-tailed percentile bootstrap test to determine significance at $\alpha=0.05$.} better than FactCC. This result is alarming because \textit{density} is a spurious correlate, yet it can achieve similar performance as the metrics that have been trained for faithfulness evaluation. 

Moreover, we also see that both FactCC and DAE have a significantly higher correlation with \textit{density} than they do with human scores (Table \ref{tab:sum_metric_correlate}). This indicates that these metrics may rely upon spurious correlations and are not yet capturing a deeper understanding of faithfulness.

\begin{table*}[h] 
\begin{center}
\begin{tabular}{|c|l|rrrr|}
\hline
 &  &  Human  &  Perplexity & Length & PPL+Len  \\ 
\hline
\multirow{3}{*}{PersonaChat} 
& DialogRPT  & -0.033  & -0.017  & \textbf{0.086} & 0.068 \\
&  Maude  & 0.303 & \textbf{0.373} & -0.089 & 0.137 \\
& USL-H  & 0.496 & 0.092 & \textbf{0.506} & 0.469 \\  
\hline
\hline
\multirow{3}{*}{TopicalChat} 
& DialogRPT  & 0.117 & -0.011 & 0.272 & \textbf{0.276} \\
&  Maude  & 0.135 & \textbf{0.243} & -0.191 & -0.148 \\
& USL-H  & 0.318 & 0.037 & \textbf{0.359} & 0.355 \\ 
\hline
\hline
\multirow{3}{*}{DailyDialog} 
& DialogRPT & 0.025 & -0.182  & \textbf{0.359}  &  0.270  \\
&  Maude  & -0.074 & -0.076 & \textbf{0.102} & 0.033 \\
& USL-H  & 0.094 & 0.048 & -0.208 & \textbf{-0.236} \\ 
\hline
\end{tabular}
\end{center}
\caption{Correlation of the metrics with human scores and spurious correlates. Reference-free evaluation metrics have higher correlation with spurious correlates than the human scores.}
\label{tab:dialog_metric_corr}
\end{table*}

\subsection{Learned Metrics for Dialog Generation}

Dialog generation systems need to be able to generate a response given the dialog context. The ability to automatically evaluate the quality of a response is essential for building dialogue systems. \newcite{liu-etal-2016-evaluate} show that referenced-based evaluation metrics do not correlate well with human judgments of response quality. This has led to an increased interest in reference-free evaluation metrics for evaluating dialogue response quality. 

Similar to our analysis in \cref{section_faithfulness_summarization}, we aim to look at recently proposed metrics for reference-free evaluation, along with spurious correlates for dialog response quality, and compare them against human judgments. 

\textbf{DialogRPT.} \newcite{gao-etal-2020-dialogue} finetune GPT-2 to predict the different types of human feedback (replies, upvotes, etc.) in Reddit threads and combine these to form a composite score for response quality. They evaluate their approach on the Reddit data that they collected and show that their method achieves higher example-level agreement with human judgments than baseline metrics.

\textbf{MAUDE.} \newcite{sinha-etal-2020-learning} propose a model that encodes each utterance in the dialog context using a pre-trained BERT model and leverages the temporal transitions between them to score a response. They add noise to existing dialog responses to create negative examples and train their system to distinguish them from valid responses using noise contrastive estimation (NCE). They evaluate their model on the PersonaChat \cite{zhang-etal-2018-personalizing} dataset and report improved example-level Spearman correlation with human judgments compared to existing baseline metrics.

\textbf{USL-H.} \newcite{phy-etal-2020-deconstruct} decompose response quality into three aspects and train a model to score a response along each of these aspects. They then combine the scores hierarchically into one composite score for response quality. They evaluate their metric on the DailyDialog \cite{li-etal-2017-dailydialog} dataset and report significantly higher example-level correlations than previous baseline metrics.

\textbf{MNLI+Adv.} \newcite{dziri2021evaluating} introduce an entailment-based metric that evaluates the groundedness of a dialog response, i.e., whether the generated response is consistent with the information in the provided external context, such as a Wikipedia article. They trained their metric on automatically generated adversarial data by applying perturbations to the evidence. They further collect human annotations for the various aspects of dialog generation, such as entailment, genericness, etc., and show that their method is more effective in accurately categorizing the generations than existing entailment models.

To assess these metrics, we look at two spurious correlates for dialog quality -- perplexity and length of the generated output -- as well as a simple combination of two measures. We compute perplexity using a pre-trained GPT-2 language model \cite{radford2019language}. Perplexity (PPL) and length are spurious correlates since they do not account for the dialog context, and therefore it is possible to have high-quality and low-quality responses with similar perplexities/lengths. For groundedness evaluation, we look at the same word overlap measures, as we did for summarization, i.e., \textit{density} and \textit{coverage}, and we measure overlap between the response and the provided external evidence.

\textbf{Results.} We evaluate metrics\footnote{We use the code provided by \newcite{yeh-etal-2021-comprehensive} for these experiments.} for response quality estimation on three popular multi-turn dialog datasets -- DailyDialog, which contains dialogs about everyday topics \cite{li-etal-2017-dailydialog}, TopicalChat, which contains dialogs conditioned on a set of $8$ broad topics \cite{Gopalakrishnan2019}, and PersonaChat, which contains dialogs conditioned on personas \cite{zhang-etal-2018-personalizing}.

To evaluate the recently proposed metric for response groundedness, we use human annotations collected by \newcite{dziri2021evaluating} on Wizard of Wikipedia \cite{dinan2019wizard}, a dataset that consists of dialogues conditioned on information from Wikipedia articles. In particular, we use their entailment annotations, where human annotators judge whether or not the external evidence entails a generated response.

Figure \ref{fig:dialog_metric_human} shows the correlations with the human scores and the spurious correlates for the dialog generation evaluation metrics. In DialyDialog, we find that perplexity achieves a similar correlation with human judgments as USL-H. In TopicalChat, perplexity or length alone does not beat out any of the learned metrics; however, combining the two measures achieves a significantly better correlation with humans than learned metrics. In PersonaChat, USL-H achieves the highest correlation with human judgment, though the combined PPL+Len score is close. We observe that USL-H is more consistent than the other reference-free metrics and achieves significantly higher correlations with human scores than MAUDE and DialogRPT for PersonaChat and TopicalChat. We further find that the reference-free metrics have a higher correlation with the spurious correlates than the human scores (Table \ref{tab:dialog_metric_corr}), which again suggests that these learned metrics may be relying upon spurious correlations.

\begin{figure}[h]
\centering
\includegraphics[width=1.1\linewidth]{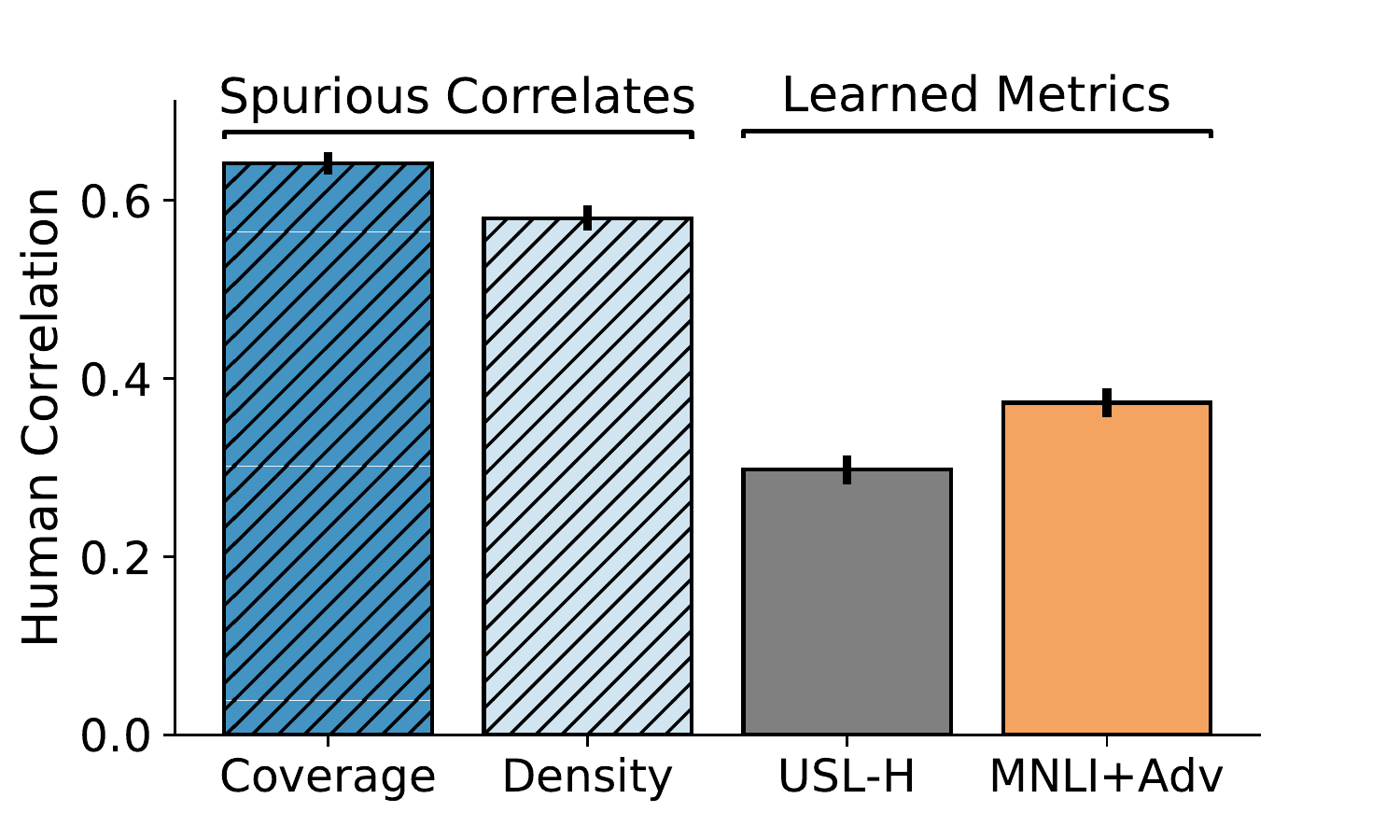}  
  \caption{Correlation of the spurious correlates and learned metrics with human scores on groundedness evaluation. Both coverage and density get significantly higher correlations with human scores than the learned metrics.}
  \label{fig:begin_human}
\end{figure}

For groundedness evaluation\footnote{We do not include MAUDE and DialogRPT results for this task since they perform significantly worse.}, both \textit{coverage} and \textit{density} achieve significantly higher correlation with human scores than MNLI+Ad and USL-H. Furthermore,  MNLI+Ad and USL-H get a higher correlation with these spurious correlates than human scores (Figure \ref{fig:begin_human}).

Despite relatively high correlations on their original datasets, these metrics seem to perform similarly to simple spurious correlations on other datasets. In order to better understand the effectiveness of these reference-free evaluation metrics, we suggest that future research includes comparisons to potential spurious correlates and that research communities come up with a set of potential standard spurious correlates.

\begin{table}[]
    \centering
    \begin{tabular}{|l|c|c|c|}
    \hline
      Metric & Human & Coverage & Density \\ 
      \hline
      USL-H  & 0.298 & 0.467 & \textbf{0.515} \\ 
      MNLI+Adv & 0.373 & 0.451  & \textbf{0.514} \\   
      \hline
    \end{tabular}
    \caption{Correlation of USL-H  and MNLI+Adv scores with humans vs coverage and density. Both learned metrics have a significantly higher correlation with density than human scores.}
    \label{tab:begin_metric_correlate}
\end{table}

\section{Learned Metrics in System-level Evaluation} \label{sec:system_level}

\begin{figure*}[t]
\centering
\makebox[0pt]{\includegraphics[scale=0.58]{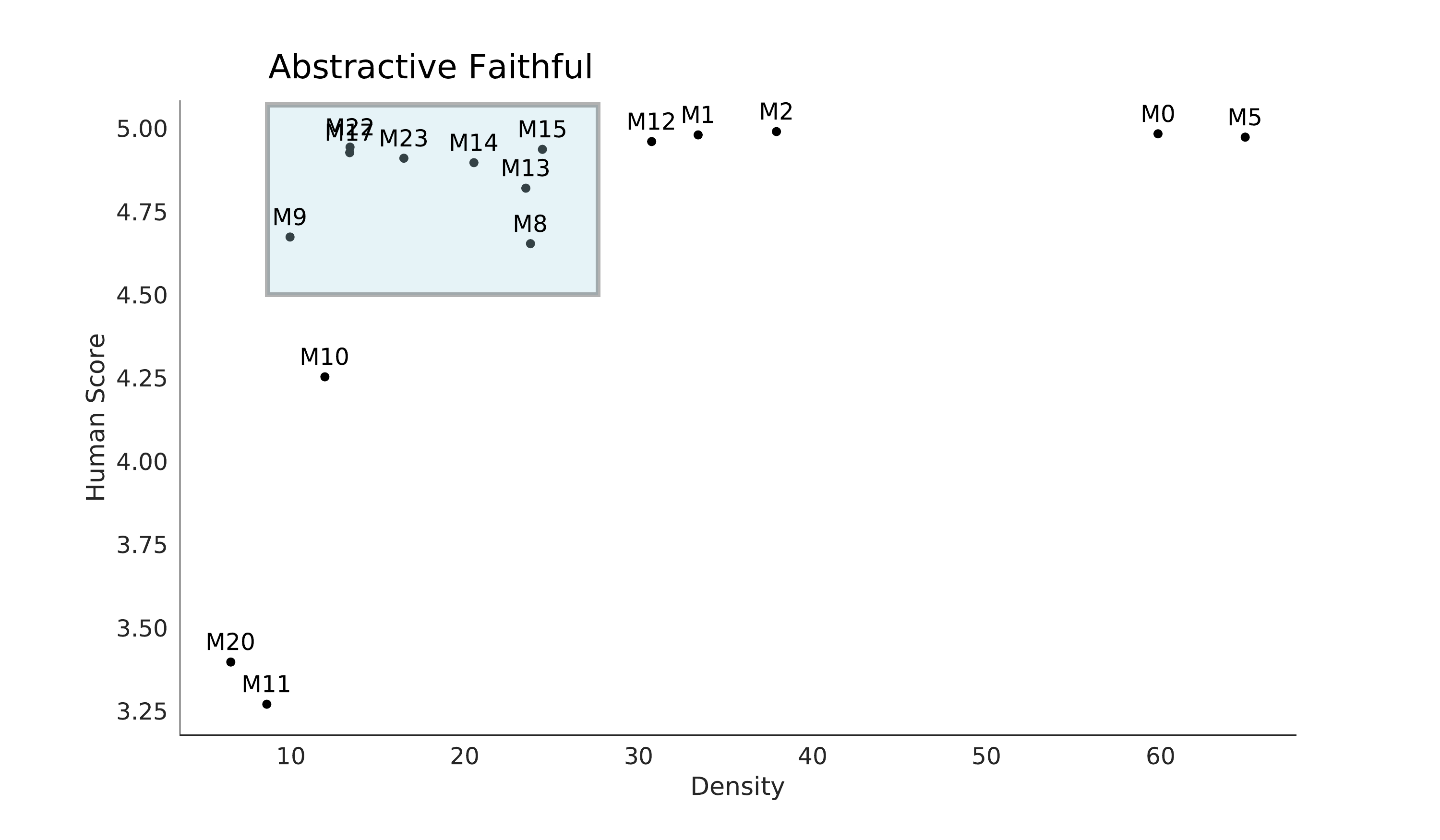}}
\caption{Density and human scores for summarization systems. We analyze the accuracy of the metrics in ranking all the systems vs. ranking the systems within abstractive faithful group, shown in the blue box. Abstractive faithful systems have faithfulness score higher than 4.5 (out of 5) and density lower than 30.}
\label{fig:density_vs_human_score}
\end{figure*}

\subsection{Pairwise Ranking of Systems}

Our example-level analysis demonstrates that recently proposed learned evaluation metrics achieve worse correlations with human scores than spurious correlates for almost all the settings. Since an important goal of building these metrics is to be able to rank arbitrary systems, we analyze whether these concerns we observe at the example level manifest into harms at the system level (i.e., ranking systems incorrectly). In order to study this, we need a large collection of human evaluation data across a wide range of systems. \newcite{fabbri2020summeval} have recently released human evaluations for faithfulness across $16$ summarization systems on CNN/DM. Therefore, we focus on system-level rankings of faithfulness for the remainder of the paper.

We first measure pairwise ranking accuracy for all the systems shown in \autoref{fig:density_vs_human_score}.\footnote{Citations corresponding to these systems are included in Appendix \ref{appendix}.} We find that system-level rankings suffer from a similar issue as the example level correlations: density and coverage appear as spurious correlations (Table \ref{tab:overall_accuracy}). From this observation, we perform a finer-grained analysis and show that these factuality metrics fail on the most important subset of model comparisons: abstractive but faithful summarization system (AF) -- where the current state-of-the-art abstractive summarization systems fall.

\begin{table}[]
\centering
\begin{tabular}{|l|c|c|}
\hline
        & All Pairs & Within AF\\ \hline
Coverage & 56.54    &    26.60                   \\
Density  & \textbf{81.01}    &    \textbf{40.45}               \\ 
FactCC   & 78.87    &    38.26                 \\ 
DAE      & 80.39    &    37.88                \\ \hline
\end{tabular}
\caption{Accuracy of pairwise ranking across all the systems and within Abstractive Faithful (AF). We observe that the ranking accuracy of all metrics is significantly lower for systems within AF compared to all pairs. Density performs as well as the best learned metric (DAE) in both cases.}
\label{tab:overall_accuracy}
\end{table}

\begin{figure*}[ht]
\centering
\makebox[0pt]{\includegraphics[scale=0.56]{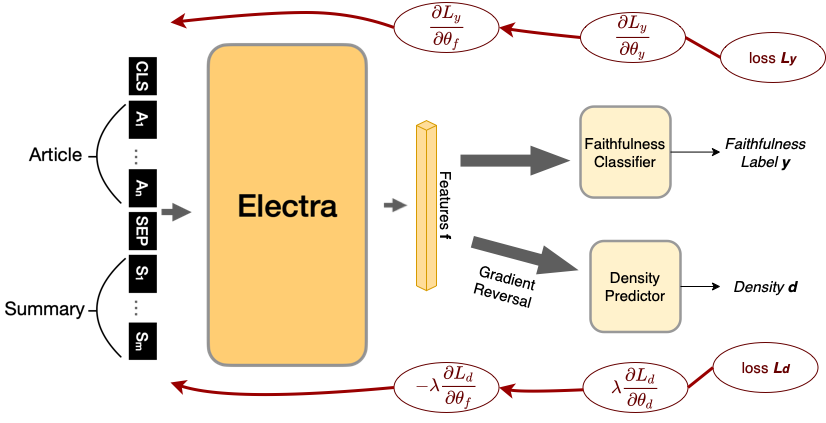}}
\caption{Architecture of adversarial model. The input sequence is first encoded via a pre-trained Electra model, and the representation is used for both faithfulness classification and density prediction. Gradients from the density predictor are reversed in order to make updates to the encoder's parameters, forcing the model to learn representations that are not predicitve of density.}
\label{fig:adversarial_model}
\end{figure*}

\subsection{Results} 
Both faithfulness metrics perform relatively well when we look at pairwise ranking accuracy across all pairs of models (Table \ref{tab:overall_accuracy}). However, they are unable to improve over \textit{density}, which achieves the highest overall accuracy. When we look at ranking within the abstractive faithful group, we see \textit{density} is no longer a good measure for the faithfulness of a system since these systems are relatively close in terms of density. Similarly, the performance of the learned metrics drops significantly, which is an expected result since our analysis in \cref{section_faithfulness_summarization} showed that both FactCC and DAE are spuriously correlated with density. We claim that our system-level analysis is further evidence that these metrics may be relying heavily on simple spurious measures such as word overlap.

These results highlight the importance of performing analyses across different distributions of systems. If we were looking at just the overall ranking accuracy of the metrics, we would conclude that DAE and FactCC correctly measure faithfulness. However, on closer examination, we see that both metrics perform relatively poorly in ranking AF systems, which is arguably the most crucial group since most state-of-the-art systems operate in this regime, and there is substantial interest in building abstractive and faithful summarization systems.

\begin{table}[h]
\centering
\begin{tabular}{|l|c|c|}
\hline
&  All Pairs & Within AF    \\ 
        \hline
FactCC-Electra & 77.85 & 27.70  \\ 
FactCC & 78.87 & 38.26 \\ 
DAE    &    80.39   & 37.88   \\ 
\hline 
\hline
Adversarial  &  \textbf{85.27}  & \textbf{59.20}  \\ \hline
\end{tabular}
\caption{Pairwise ranking accuracy for systems across All Pairs vs. Within Abstractive Faithful (AF) for DAE and Adversarial. Adversarially trained metric performs significantly better for the systems within AF than previously proposed metrics.}
\label{tab:adversarial_result}
\end{table}

\section{Adversarial Model}
In our earlier example-level analysis, we found that learned metrics have higher correlation with spurious correlates than human judgment. We further saw in our system-level analysis that learned metrics for faithfulness are unable to outperform density. One natural question that follows is whether we can build metrics that do well at the systems level by learning representations that rely less on spurious correlates. 

In order to do this, we train an entailment based model using the synthetically generated data from FactCC in an adversarial setup similar to \newcite{ganin2016domain}. In particular, our approach augments the standard faithfulness predictor with a density predictor that tries to predict the density of the summary from the model's internal representation. We use this density predictor as an adversary, and our goal is to predict faithfulness while ensuring that it is difficult to predict density using this same representation. To achieve this, the gradients from the density predictor are reversed, which makes it harder to predict the density from the encoder's representation, and thus makes the faithfulness predictions less reliant on density. The model architecture is shown in Figure \ref{fig:adversarial_model}. We initialize the parameter $\lambda$ to $0$ and gradually increase it to $1$, following the schedule detailed in \newcite{ganin2016domain}.

We fine-tune a pre-trained Electra model \cite{clark2020electra} using the transformers library \cite{wolf-etal-2020-transformers} for this task. We chose Electra in order to match the model architecture in DAE. Since the original FactCC metric was fine-tuned on BERT, we also fine-tune our own version of FactCC on Electra (FactCC-Electra) as an ablation. Our adversarially trained model is essentially the same as FactCC-Electra, but with an additional adversarial head for predicting density.

\textbf{Results.} We note that the FactCC-Electra model performs worse than the original FactCC, which is consistent with the findings in \newcite{goyal-durrett-2021-annotating}. Our adversarially trained metric has a significantly lower example-level correlation with density (27.71\%), as compared to FactCC (59.10\%) and DAE (76.37\%). We find that the adversarial model\footnote{Our adversarially trained model can be found at \href{https://github.com/esdurmus/adversarial_eval}{https://github.com/esdurmus/adversarial\_eval}.} can achieve a significantly better performance than existing learned evaluation metrics in ranking systems within the abstractive faithful (AF) group (Table \ref{fig:adversarial_model}). This suggests that it is possible to learn effective metrics that are not overly reliant on spurious correlates. Furthermore, our metric is also effective in overall pairwise ranking of the systems achieving $85.27\%$ accuracy.
\section{Related Work}

Most existing work on assessing the evaluation methodology of evaluation metrics has focused on reference-based evaluation. For example, \newcite{mathur-etal-2020-tangled} take a critical look at the use of example-level correlations to measure reference-based evaluation metrics in Machine Translation. They show that evaluating these metrics using example-level correlations can be sensitive to the presence of outliers which can lead to false conclusions about a metric's efficacy. Furthermore, \newcite{DBLP:journals/corr/abs-2107-10821} show that proper assessment of evaluation metrics is crucial as uninformed use of automated metrics such as BLEU can lead to bad deployment decisions. \newcite{caglayan-etal-2020-curious} has shown that automated reference-based evaluation metrics have robustness issues which can cause them to score generated outputs higher than human written outputs. Furthermore, \newcite{bhandari-etal-2020-evaluating} has studied the limitations of reference-based evaluation metrics of text summarization, comparing these metrics across different datasets and application scenarios. In contrast, our work focuses on analyzing learned, reference-free evaluation metrics in summarization and dialog generation, accounting for potential spurious correlates for these evaluation tasks. 

There has been some recent work comparing existing reference-free evaluation metrics for text summarization and dialog generation. \newcite{pagnoni-etal-2021-understanding} has measured the efficacy of existing reference-free faithfulness evaluation metrics of summarization on two different summarization datasets relying on example-level correlations. Similarly, \newcite{gehrmann-etal-2021-gem} has evaluated automated metrics of text summarization across a wide range of datasets.  \newcite{gabriel-etal-2021-go} has proposed a meta-evaluation framework to evaluate the evaluation metrics looking at certain aspects of these metrics such as robustness, sensitivity, high correlation with human scores, etc., and measured existing evaluation metrics across these aspects. \newcite{yeh-etal-2021-comprehensive} perform a comprehensive study of existing dialog generation metrics across several different datasets and find that the performance of metrics varies widely across datasets. 

\newcite{gabriel-etal-2021-go} and \newcite{yeh-etal-2021-comprehensive} are the most related to our work since they study robustness of these metrics looking at their performance across different datasets. In our work, however, we explicitly study spurious correlations and show that these may potentially be contributing to the robustness issues. We further present initial promising results suggesting that controlling for these spurious correlates may result in more robust evaluation metrics.

\section{Conclusion}
In conclusion, we study reference-free evaluation metrics for summarization and dialog generation and show that simply looking at overall example-level correlation with human judgment paints an incomplete picture of the effectiveness of a metric. In particular, we show that these metrics are unable to do better than simple spurious correlates for the task. We see that this trend carries over in system-level ranking for summarization systems, where a spurious correlate for the task performs as well as existing learned evaluation metrics. We find that despite the relatively high overall system-level ranking performance, the learned metrics are not robust to distribution shifts. We show that they fail to properly rank abstractive and (relatively) faithful systems, which is where the current state of the art operates. Finally, we train a faithfulness metric that scores the faithfulness of a summary without relying on the spurious overlap correlate. We show that our metric is more robust across distribution shifts and does better at ranking abstractive, faithful summarization systems. 

We suggest that future work in designing reference-free evaluation metrics should be mindful of the distribution of the evaluation data. In particular, metrics should be assessed across different distributions of systems in order to test for robustness and failure modes. Simple spurious correlates can be used as a tool to indicate potential overestimates of the effectiveness of proposed metrics. Finally, we highlight the importance of collecting large-scale human evaluation datasets across a wide range of systems, similar to \newcite{fabbri2020summeval}, to enable more comprehensive analyses of evaluation metrics.

\section{Acknowledgements}
ED is supported by SAIL Postdoc Fellowship. We further thank the anonymous reviewers and the Stanford NLP group for their invaluable feedback.
\bibliography{anthology,custom}
\bibliographystyle{acl_natbib}

\clearpage
\appendix
\section{Text Summarization Models} \label{appendix}

\begin{table}[h]
\centering
\begin{tabular}{|l|l|}
\hline
  Model Name & Paper    \\ 
 \hline 
 M0 &  Lead-3 baseline \\ 
 M1 & \newcite{zhou-etal-2018-neural-document} \\ 
 M2 & \newcite{dong-etal-2018-banditsum} \\ 
 M5 & \newcite{DBLP:conf/aaai/WuH18} \\ 
 M8 & \newcite{see-etal-2017-get} \\ 
 M9 & \newcite{chen-bansal-2018-fast} \\ 
 M10 & \newcite{gehrmann-etal-2018-bottom} \\ 
 M11 & \newcite{kryscinski-etal-2018-improving} \\
M12 & \newcite{hsu-etal-2018-unified} \\
M13 & \newcite{pasunuru-bansal-2018-multi} \\
M14 & \newcite{guo-etal-2018-soft} \\
M15 & \newcite{jiang-bansal-2018-closed} \\
M17 & \newcite{DBLP:journals/corr/abs-1910-10683} \\
M20 & \newcite{ziegler2019finetuning} \\
M22 & \newcite{lewis-etal-2020-bart} \\
M23 & \newcite{Zhang2020PEGASUSPW} \\
\hline
\end{tabular}
\caption{Models that are used in \cref{sec:system_level}.}
\label{tab:model_name_paper}
\end{table}
\end{document}